\documentclass[letterpaper]{article} 
\usepackage{aaai2026}  
\usepackage{times}  
\usepackage{helvet}  
\usepackage{courier}  
\usepackage[hyphens]{url}  
\usepackage{graphicx} 
\usepackage{natbib}  
\usepackage{caption} 
\usepackage{algorithm}
\usepackage{algorithmic}

\usepackage{newfloat}
\usepackage{listings}
\DeclareCaptionStyle{ruled}{labelfont=normalfont,labelsep=colon,strut=off} 
\floatstyle{ruled}
\newfloat{listing}{tb}{lst}{}
\floatname{listing}{Listing}
\title{Difference Vector Equalization for Robust Fine-tuning of Vision-Language Models}
\author{
    Satoshi Suzuki, Shin'ya Yamaguchi, Shoichiro Takeda, Taiga Yamane, Naoki Makishima,\\
    Naotaka Kawata, Mana Ihori, Tomohiro Tanaka, Shota Orihashi, Ryo Masumura
}
\affiliations{
    NTT, Inc., Japan\\

    satoshixv.suzuki@ntt.com
}

\usepackage{booktabs}
\usepackage {arydshln}
\usepackage{amsmath}
\usepackage{amssymb}
\usepackage{multirow}
\usepackage{bm}
\makeatletter
\newcommand{\thickhline}{%
    \noalign {\ifnum 0=`}\fi \hrule height 1pt
    \futurelet \reserved@a \@xhline
}
\usepackage{hhline}

\usepackage{bibentry}

\begin{document}

\maketitle

\begin{abstract}
Contrastive pre-trained vision-language models, such as CLIP, demonstrate strong generalization abilities in zero-shot classification by leveraging embeddings extracted from image and text encoders.
This paper aims to robustly fine-tune these vision-language models on in-distribution~(ID) data without compromising their generalization abilities in out-of-distribution~(OOD) and zero-shot settings.
Current robust fine-tuning methods tackle this challenge by reusing contrastive learning, which was used in pre-training, for fine-tuning.
However, we found that these methods distort the geometric structure of the embeddings, which plays a crucial role in the generalization of vision-language models, resulting in limited OOD and zero-shot performance.
To address this, we propose \textbf{Difference Vector Equalization}~(DiVE), which preserves the geometric structure during fine-tuning.
The idea behind DiVE is to constrain \textit{difference vectors}, each of which is obtained by subtracting the embeddings extracted from the pre-trained and fine-tuning models for the same data sample.
By constraining the difference vectors to be equal across various data samples, we effectively preserve the geometric structure.
Therefore, we introduce two losses: average vector loss~(AVL) and pairwise vector loss~(PVL).
AVL preserves the geometric structure globally by constraining difference vectors to be equal to their weighted average.
PVL preserves the geometric structure locally by ensuring a consistent multimodal alignment.
Our experiments demonstrate that DiVE effectively preserves the geometric structure, achieving strong results across ID, OOD, and zero-shot metrics.
\end{abstract}


\section{Introduction}
\label{sect:intro}

Advances in contrastive pre-training have led to the development of vision-language models, such as CLIP~\cite{Radford21} and ALIGN~\cite{Jia21}.
These models consist of two separate encoders for visual and textual modalities.
They are pre-trained on large-scale image-text datasets to learn multimodal representations by mapping images and texts into a shared embedding space.
After pre-training, the model can execute classification by using these embeddings even for previously unseen tasks.
Specifically, it prepares text prompts for possible class names and selects the class, the text embedding of which has the highest similarity to the image embedding.
Such ``zero-shot'' classification using vision-language models achieves reasonable performance on many tasks~\cite{Radford21} and impressive robustness to various distribution shifts, such as those from natural images to sketches~\cite{Wang19_sketch}.

\begin{figure}[t]
   \centering
    \includegraphics[width = \columnwidth] {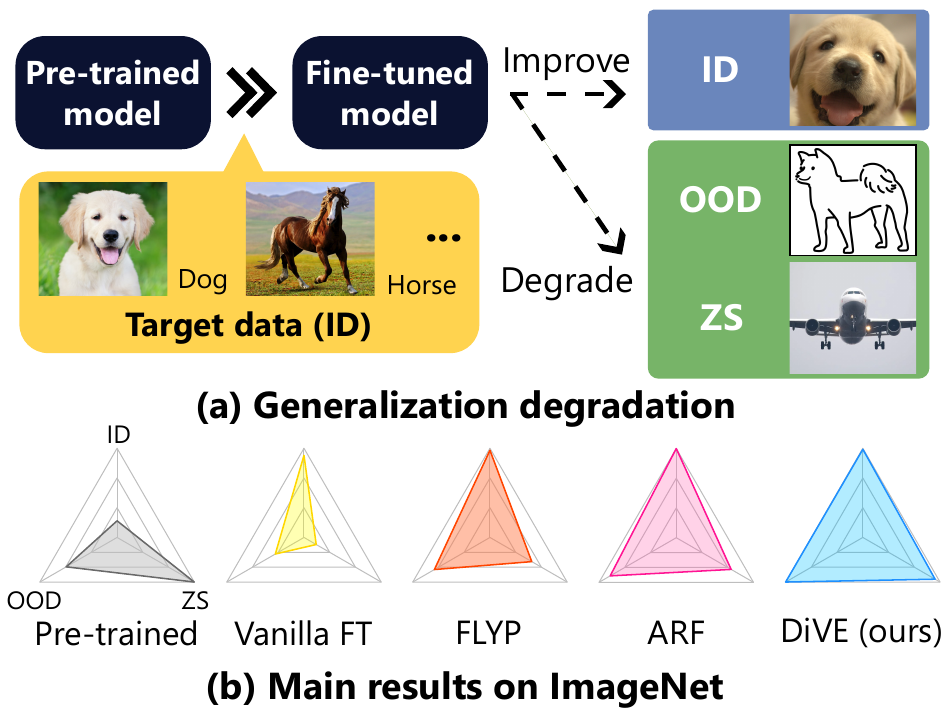}
    \vspace{-3.0mm}
    \caption{(a)~Illustrative example of generalization degradation.
    Performance of pre-trained models in out-of-distribution~(OOD) and zero-shot~(ZS) settings severely degrades after vanilla fine-tuning~(FT) on in-distribution~(ID) data.
    (b)~Normalized performance of robust fine-tuning methods on ID, OOD, and ZS metrics.
    We used ImageNet as target task.
    DiVE performs well across all metrics.
    }
  \label{fig:radar}
\end{figure}

Despite advances in vision-language models, their zero-shot performance is often sub-optimal, especially on data samples that deviate from the pre-training domain, such as satellite images~\cite{Radford21}.
In such cases, task-specific fine-tuning is commonly used to enhance performance.
However, fine-tuning a pre-trained model can easily degrade its generalization ability.
Specifically, vanilla fine-tuning~(FT) improves performance on in-distribution~(ID) target data but significantly degrades out-of-distribution~(OOD) and zero-shot performance~\cite{Kumar22,Ilharco22}.
Figure~\ref{fig:radar}~(a) illustrates an example: while vanilla FT improves classification performance on ID data samples~(e.g., dog), it degrades OOD performance on samples from the same classes but in different styles~(e.g., sketch), as well as zero-shot performance on samples from classes not seen during fine-tuning~(e.g., airplane).
Given this phenomenon, our goal is to maximize ID performance without compromising generalization ability.
Achieving this could enhance the practical value of vision-language models in diverse and dynamic environments such as self-driving cars.

To tackle this challenge, robust fine-tuning methods for vision-language models have been proposed.
For example, FLYP~\cite{Goyal23} executes contrastive learning during fine-tuning using class label prompts~(e.g., ``a photo of a dog''), unlike vanilla FT, which updates only the image encoder~\cite{Wortsman22}.
This strategy is expected to enhance the generalization ability by considering both image and text encoders.
Subsequently, ARF~\cite{Han24} introduces an auxiliary contrastive loss into FLYP using a reference image-caption dataset with diverse semantics.
This approach works as a replay technique~\cite{Rolnick19}, retaining pre-trained knowledge.

While these robust fine-tuning methods demonstrate promising performance, there may still be room for improvement in both OOD and zero-shot performance.
As one interesting fact, we found that these methods distort the geometric structure of the embeddings learned by pre-training~(see Table~\ref{tab:rsa}), i.e., the relative positions between the embeddings.
Since the relative distances between embeddings reflect the semantic similarity of corresponding inputs, this structure is fundamental to the strong generalization ability of vision-language models~\cite{Goel22,Oh23,Yamaguchi25}.
Consequently, its distortion can degrade both OOD and zero-shot performance.
Therefore, the following research question naturally arises: \textit{Can we improve robust fine-tuning for vision-language models by preserving the geometric structure of the embeddings?}

\begin{figure*}[t]
   \centering
    \includegraphics[width = 150mm] {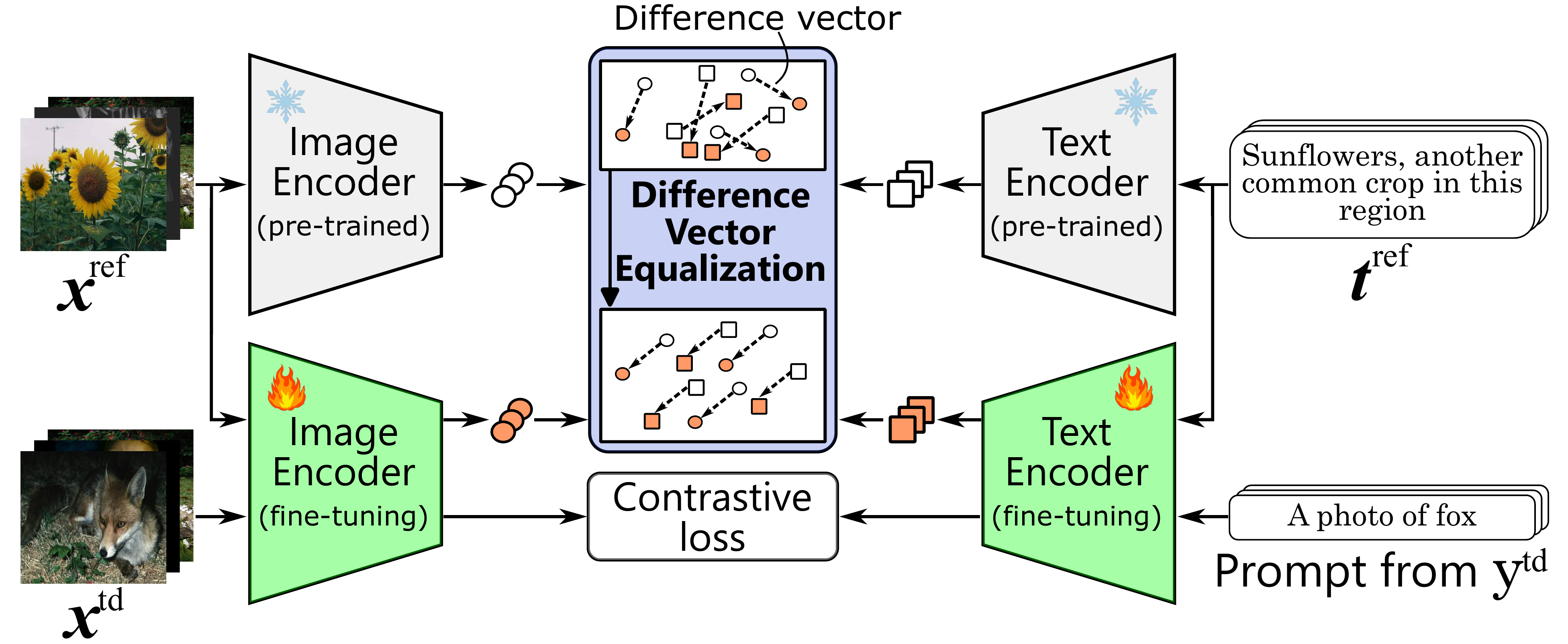}
    \caption{Overview of our proposed method, \textbf{Difference Vector Equalization}~(DiVE).
    It uses contrastive loss for fine-tuning.
    While fine-tuning on target data, it constrains all difference vectors for reference data~($\bm{x}^{\rm ref}$ and $\bm{t}^{\rm ref}$) to be equal.}
    \vspace{-1.0mm}
  \label{fig:prop}
\end{figure*}

In this paper, we propose a novel robust fine-tuning method \textbf{Difference Vector Equalization}~(DiVE).
DiVE builds on FLYP, i.e., using contrastive loss for fine-tuning, and introduces a constraint to preserve the geometric structure of the embeddings during fine-tuning on the target data.
The idea behind DiVE is to constrain \textit{difference vectors}, each of which is obtained by subtracting the embeddings extracted from the pre-trained and fine-tuning models for the same data sample.
Intuitively, difference vectors represent the changes in the samples in the embedding space caused by fine-tuning; thus, constraining difference vectors to be equal across various data samples can help preserve the geometric structure~(see Fig.~\ref{fig:prop}).
Therefore, we first prepare the image-caption reference dataset, which is also used with ARF, to obtain the difference vectors for various images and captions.
We then introduce two new losses for DiVE: \textit{average vector loss}~(AVL) and \textit{pairwise vector loss}~(PVL).
AVL computes a weighted average of the difference vectors and constrains each difference vector to be equal to this average vector.
Therefore, it preserves the global geometric structure formed by various image and text embeddings.
On the other hand, PVL constrains the difference vector for each reference image to be equal to that for its corresponding caption.
It preserves the local geometric structure by ensuring consistent image-caption alignment for each pair.
By incorporating AVL and PVL into FLYP, DiVE can effectively preserve the geometric structure of embeddings during fine-tuning.

We empirically demonstrate that DiVE effectively preserves the geometric structure of the embeddings learned by pre-training, as expected~(see Table~\ref{tab:rsa}).
Concurrently, DiVE performs impressively on various classification tasks.
For example, Fig.~\ref{fig:radar}~(b) shows that it consistently performs well across ID, OOD, and zero-shot metrics.
These results highlight the importance of preserving the geometric structure for robust fine-tuning.

\section{Related Work}
\label{sect:relate}

\textbf{Robust Fine-tuning for vision-language models.}
Vision-language models, such as CLIP~\cite{Radford21}, demonstrate strong generalization abilities in OOD and zero-shot settings.
However, empirical evidence suggests that vanilla FT, which updates the weights of the image encoder using downstream data, compromises these strengths~\cite{Wortsman22,Ilharco22}.
To tackle this challenge, robust fine-tuning methods for vision-language models have been proposed.
An early baseline method, LP-FT~\cite{Kumar22}, first applies linear probing then fine-tunes the entire weights to enhance OOD performance.
FLYP~\cite{Goyal23} uses vision-language contrastive learning for fine-tuning.
This choice can enhance OOD performance over vanilla FT by considering both image and text encoders.
ARF~\cite{Han24} is the first robust fine-tuning method that explicitly aims to enhance not only OOD but also zero-shot performance.
It combines FLYP and a replay technique~\cite{Rolnick19}, which involves training on a reference dataset~(e.g., CC3M~\cite{Sharma18_cc3m}) that resembles the pre-training data, to help recover the inherent generalization ability of the pre-trained model.
While FLYP and ARF show promising performance, both OOD and zero-shot performance can still be improved.
DiVE improves them by introducing a new perspective: preserving the geometric structure of the embeddings.

\noindent
\textbf{Geometry of Embeddings of Contrastive Learning Models.}
Contrastive learning has been widely used for representation learning on various modalities~\cite{Chen20_simclr,Gao21}.
Most contrastive learning methods adopt $L_2$-normalized embeddings for stable learning~\cite{Xu18}, which causes the embeddings to lie on a unit hypersphere.
\citet{Wang20_contrastive} found that the relative positions between hyperspherical embeddings, i.e., the geometric structure, play a crucial role in generalization in contrastive learning models.
Studies~\cite{Goel22,Oh23,Yamaguchi25} have shown that this structure is also crucial for contrastive pre-trained vision-language models, such as CLIP.
Among these, CyCLIP~\cite{Goel22} shares some concepts with DiVE in constraining the geometric structure of the embeddings.
Specifically, it constrains the cosine similarities between in-modal and cross-modal embeddings during pre-training.
We found that such a cosine similarity-based constraint is insufficient to preserve the geometric structure and improve performance, as it captures only the angular relationship between embeddings.
In contrast, our difference vector-based constraint achieves both~(see Table~\ref{table:cyclip}).

\noindent
\textbf{Maintaining Zero-shot Performance.}
There are several approaches for maintaining the zero-shot performance of vision-language models during fine-tuning, not only robust fine-tuning.
Prompt learning~\cite{Zhou22_coop,Zhou22_cocoop,Khattak23} optimizes a set of learnable prompt vectors, while freezing the weights of the pre-trained model.
This strategy perfectly maintains the generalization ability.
However, as reported by \cite{Shu23}, its classification performance on target data is often limited due to the small number of learnable parameters.
Methods for continual learning~\cite{Wang24} can be adapted to our setting by interpreting the contrastive pre-training as the previous task.
For example, the state-of-the-art method SnD~\cite{Yu24} is related to DiVE.
Specifically, it leverages a reference dataset~(e.g., CC3M) for feature distillation, i.e., constraining the image embeddings from the pre-trained and fine-tuning models to be identical during continual learning.
In our context, this can be viewed as forcing the difference vectors to be zero.
However, such a strong constraint may hinder adaptation to target data.
In contrast, DiVE allows for non-zero difference vectors, offering greater flexibility and better performance~(see Tables~\ref{tab:comp}, \ref{tab:zero}, and \ref{tab:iwild_fmow}).

\noindent
\textbf{Weight Ensemble.}
In an orthogonal line of research, several studies~\cite{Wortsman22,Ilharco22} demonstrated that ensembling the weights of pre-trained and fine-tuned models can improve OOD and zero-shot performance.
Although they were originally proposed in integration with vanilla FT, DiVE can also benefit from such ensembling strategies.
We present the results of integrating DiVE into this strategy in Sec.~\ref{sec:app_wise} in the Appendix.

\section{Preliminaries}

We fine-tune a contrastive pre-trained vision-language model, such as CLIP~\cite{Radford21}, on target data. 
A vision-language model comprises an image encoder $f_\theta: \mathcal{X}\to\mathbb{R}^d$ and text encoder $g_\phi: \mathcal{T}\to\mathbb{R}^d$, where $\theta$ and $\phi$ represent the weights of the image and text encoders, respectively.
Furthermore, $\mathcal{X}$ and $\mathcal{T}$ denote the image and text spaces, respectively, and $d$ is the dimension of the embeddings.
During the pre-training, image-text pairs are sampled from a web-scale training dataset $\mathcal{S}=\{(\bm{x}_i, \bm{t}_i) \}^{N}_{i=1}$, where $\bm{x}_i \in \mathcal{X}$ is the $i$-th image sample and $\bm{t}_i \in \mathcal{T}$ is the corresponding text description.
A contrastive loss function $\mathcal{L}_{\rm cl}$ is used to align the image embedding $f_\theta(\bm{x}_i)$ with the corresponding text embedding $g_\phi(\bm{t}_i)$,
\begin{align}
\centering
    \mathcal{L}_{\rm cl} = &-\frac{1}{2B}\sum^B_{i=1}\log
    \frac{\exp(f_{\theta}(\bm{x}_i) \cdot g_{\phi}(\bm{t}_i)/\tau)}
    {\sum^B_{k=1}\exp(f_{\theta}(\bm{x}_i) \cdot g_{\phi}(\bm{t}_k)/\tau)} \nonumber \\
    &-\frac{1}{2B}\sum^B_{i=1}\log
    \frac{\exp(f_{\theta}(\bm{x}_i) \cdot g_{\phi}(\bm{t}_i)/\tau)}
    {\sum^B_{k=1}\exp(f_{\theta}(\bm{x}_k) \cdot g_{\phi}(\bm{t}_i)/\tau)}.
\label{formula:cl}
\end{align}
Here, $B$ is the batch size, $\tau$ denotes the temperature used to scale the pairwise similarities, and $\cdot$ indicates an inner product.
Both encoders output $L_2$-normalized embeddings, making their inner product equivalent to cosine similarity.

During inference, we first prepare text prompts $\bm{t}^c \in \mathcal{T}$ for each class of interest $c \in \mathcal{C}$, such as
\begin{align}
\centering
    \bm{t}^c = \text{``a photo of a [$c$]''},
\label{formula:prompt}
\end{align}
where [$c$] represents the $c$-th class name.
Even for previously unseen tasks, i.e., the zero-shot setting, the model classifies a test image $\bm{x}^{\rm test}$ by selecting the class $\hat{y}$ whose $\bm{t}^c$ has the highest similarity to the image embedding,
\begin{align}
\centering
     \hat{y}= \arg \max_{c\in\mathcal{C}}~f_\theta(\bm{x}^{\rm test}) \cdot g_\phi(\bm{t}^c).
\label{formula:zero}
\end{align}

We consider a fine-tuning problem that updates the weights of encoders, i.e., $\theta$ and $\phi$, using the target data $\mathcal{S}^{\rm td}=\{(\bm{x}_i^{\rm td}, y_i^{\rm td})\}^{N_{\rm td}}_{i=1}$.
With DiVE, we inherit the basic fine-tuning strategy of FLYP, which leverages supervision via a contrastive loss.
Namely, we use the loss function $\mathcal{L}_{\rm cl}$ in Eq.~(\ref{formula:cl}) on $\mathcal{S}^{\rm td}$, where the class label $y_i^{\rm td}$ is converted into a text prompt, as described in Eq.~(\ref{formula:prompt}).

\section{Proposed Method}

This paper proposes DiVE, which explicitly preserves the geometric structure of the embeddings.
Figure~\ref{fig:prop} shows an overview of DiVE.
Similar to ARF~\cite{Han24}, we use a reference image-caption dataset $\mathcal{S}^{\rm ref} = \{(\bm{x}^{\rm ref}_i, \bm{t}^{\rm ref}_i)\}^{N_{{\rm ref}}}_{i=1}$, which resembles the pre-training dataset $\mathcal{S}$ and covers diverse semantics.
Our main idea is to constrain the difference vectors between the embeddings extracted from the pre-trained and fine-tuning models to be equal across all images and captions in $\mathcal{S}^{\rm ref}$.
The difference vectors for $\bm{x}^{\rm ref}_i$ and $\bm{t}^{\rm ref}_i$ are defined as\footnote{Strictly, embeddings lie on a unit hypersphere, so difference vectors should account for curvature. However, we found that the norm of the difference vectors is very small~($10^{-3}$ to $10^{-4}$ on average), allowing the curvature to be ignored in practice~\cite{Lee12}.},
\begin{align}
\centering
    u(\bm{x}^{\rm ref}_i) = f_{\theta_{\rm ft}}(\bm{x}^{\rm ref}_i) &- f_{\theta_{\rm pre}}(\bm{x}^{\rm ref}_i), \\
    v(\bm{t}^{\rm ref}_i) = g_{\phi_{\rm ft}}(\bm{t}^{\rm ref}_i) &- g_{\phi_{\rm pre}}(\bm{t}^{\rm ref}_i),
\label{formula:mv}
\end{align}
where $\{\theta_{\rm pre}, \phi_{\rm pre}\}$ and $\{\theta_{\rm ft}, \phi_{\rm ft}\}$ represent the weights of the pre-trained and fine-tuning models, respectively.
In other words, our ideal state is $u(\bm{x}^{\rm ref}_1)$ $=$ $v(\bm{t}^{\rm ref}_1)$ $=$ $\cdots$ $=$ $u(\bm{x}_{N_{\rm ref}}^{\rm ref})$ $=$ $v(\bm{t}_{N_{\rm ref}}^{\rm ref})$ after fine-tuning.
Achieving this state effectively preserves the geometric structure of the embeddings extracted from the pre-trained model~(see Fig.~\ref{fig:prop}), protecting it from distortion caused by fine-tuning~\cite{Rajaee21,Kumar22}.
We thus introduce two new losses, i.e., average vector loss~(AVL) and pairwise vector loss~(PVL), which we formulate in the following sections.

\subsection{Average Vector Loss}

In AVL, we first compute an average vector $\bm{m} \in \mathbb{R}^d$ from the difference vectors for the images and captions within a mini-batch from $\mathcal{S}^{\rm ref}$.
Here, $\bm{m}$ is defined as the exponential moving average of difference vectors from previous batches,
\begin{align}
\centering
    \bm{m} = \alpha \cdot \bm{m}_{\rm p} + \frac{(1 - \alpha)}{B'} \sum^{B'}_{j=1}\left(
        \frac{u(\bm{x}^{\rm ref}_j) + v(\bm{t}^{\rm ref}_j)}{2}
    \right),
\label{formula:target}
\end{align}
where $\bm{m}_{\rm p}$ is the average vector used in the immediately previous batch, and $B'$ is the batch size of the mini-batch from $\mathcal{S}^{\rm ref}$.
$B'$ does not necessarily have to be equal to the batch size of the mini-batch from target data~(i.e., $B$ in Eq.~(\ref{formula:cl})).
Furthermore, $\alpha$ is a hyper-parameter for determining the effect of the current difference vectors on updating the average vector.
By setting a large $\alpha$~(e.g., 0.99), the average vector maintains high consistency across batches while reflecting the current trend of difference vectors.
We initialize $\bm{m}$ as a zero vector because $\bm{m}_{\rm p}$ is undefined for the first batch.

By using $\bm{m}$, AVL constrains each difference vector,
\begin{align}
\centering
    \mathcal{L}_{\rm avl} = \frac{1}{B'} \sum^{B'}_{j=1} ( \|u(\bm{x}^{\rm ref}_j) - \bm{m}\|^2 + \|v(\bm{t}^{\rm ref}_j) - \bm{m}\|^2 ).
\label{formula:tam}
\end{align}
This formulation encourages all difference vectors to be equal to the average vector and preserves the global geometric structure formed by various image and text embeddings.

\subsection{Pairwise Vector Loss}

While AVL preserves the global geometric structure, PVL preserves the local geometric structure formed by each image-caption pair.
Since distortion of this local structure directly affects inference of the vision-language model using Eq.~(\ref{formula:zero}), preserving it is critical for their generalization.
PVL thus constrains the difference vectors for the reference image $\bm{x}^{\rm ref}_j$ and that for its caption $\bm{t}^{\rm ref}_j$,
\begin{align}
\centering
    \mathcal{L}_{\rm pvl} = \frac{1}{B'}\sum^{B'}_{j=1} \|u(\bm{x}^{\rm ref}_j) - v(\bm{t}^{\rm ref}_j)\|^2.
\label{formula:pam}
\end{align}

Finally, by combining AVL and PVL into FLYP, we obtain the loss for fine-tuning the pre-trained model,
\begin{align}
\centering
    \mathcal{L}_{\rm final} = \mathcal{L}_{\rm cl} + \lambda  \cdot (\mathcal{L}_{\rm avl} + \mathcal{L}_{\rm pvl}),
\label{formula:final}
\end{align}
where $\lambda$ is the hyper-parameter for determining the effect of AVL and PVL.
Furthermore, $\mathcal{L}_{\rm cl}$ is the contrastive loss defined in Eq.~(\ref{formula:cl}) using the target data $\mathcal{S}^{\rm td}$.
Note that during inference, DiVE uses only the final fine-tuned model, incurring no additional computational cost beyond the original vision-language models.

\begin{table}[t]
    \centering
        \scalebox{1.0}{
    \begin{tabular}{c c c c c c}
        \toprule
          Method        & RSA correlation score~($\uparrow$) \\ \midrule
          FLYP          & 0.825 \\ 
          FLYP + replay & 0.850 \\ 
          SnD           & 0.847 \\ \hdashline
          FLYP + AVL    & 0.978 \\
          FLYP + PVL    & 0.976 \\
          DiVE          & \textbf{0.981} \\
    \bottomrule
    \end{tabular}
}
    \vspace{-1.0mm}
    \caption{Representation similarity analysis~(RSA) correlation scores of robust fine-tuning methods.
    ``FLYP + replay'' denotes FLYP using auxiliary contrastive loss, which serves as proxy for ARF.
    We used Flickr8K dataset for evaluation.}
    \vspace{-1.0mm}
    \label{tab:rsa}
\end{table}

\begin{table*}[t]
    \centering
        \scalebox{0.96}{
    \begin{tabular}{c | c | c c c c c | c c}
        \thickhline
          Method           & ImageNet~(ID) & ImageNet-V2 & ImageNet-R & ImageNet-A & ImageNet-Sketch & ObjectNet & OOD avg. \\ \hline
          Pre-trained      & 68.3 & 61.9 & 77.7 & 50.0 & 48.3 & 55.4 & 58.7 \\ \hdashline
          Vanilla FT       & 81.3 & 71.2 & 66.1 & 37.8 & 46.1 & 53.3 & 54.9 \\
          LP-FT            & 81.7 & 72.1 & 73.5 & 47.6 & 50.3 & \textbf{58.2} & 60.3 \\
          FLYP$^\dagger$   & 82.2 & 73.0 & 71.5 & 48.4 & 49.7 & 54.8 & 59.5 \\ 
          ARF              & \textbf{82.7} & 72.8 & 75.6 & 50.3 & 51.8 & 55.8 & 61.3 \\ 
          SnD$^\dagger$    & 82.4 & 73.2 & 74.3 & 50.0 & 51.4 & 54.5 & 60.7 \\ 
          DiVE~(ours)      & 82.5 & \textbf{73.8} & \textbf{77.3} & \textbf{54.9} & \textbf{52.9} & 56.9 & \textbf{63.2} \\
    \thickhline
    \end{tabular}
}
    \vspace{-1.0mm}
    \caption{Experimental results of ID and OOD performance using CLIP ViT-B/16.
    We report results of ID dataset, i.e., ImageNet, and five OOD datasets.
    Numbers represent top-1 accuracy.
    $\dagger$ denotes our reproduction results.
    Bold indicates best result among fine-tuned methods, i.e., excluding pre-trained model.}
    \vspace{-1.0mm}
    \label{tab:comp}
\end{table*}

\section{Experiments}
\label{sect:exp}

\subsection{Datasets}
\label{subsec:data}
We used the following benchmarks to measure ID, OOD, and zero-shot performance.

\noindent
\textbf{ImageNet}~\cite{Imagenet09} is an object-centric dataset containing 1.2M training and 50,000 validation images from 1,000 classes.
We considered ImageNet-V2~\cite{Recht19}, ImageNet-R~\cite{Hendrycks21r}, ImageNet-A~\cite{Hendrycks21a}, ImageNet-Sketch~\cite{Wang19_sketch}, and ObjectNet~\cite{Barbu19} as the OOD datasets following \cite{Goyal23,Han24}.
Following common practice~\cite{He16}, we report the performance on the ImageNet validation set as ID performance.

\noindent
\textbf{iWildCam}~\cite{Beery20} consists of images from 182 animal species.
The ID and OOD datasets differ in the camera used and factors such as background and illumination.
We used the standard splits for training, validation, and ID and OOD testing~\cite{Koh21}.
As the dataset has label imbalance, we report the macro F1-score, following \citet{Goyal23}.

\noindent
\textbf{FMoW}~\cite{Christie18} consists of satellite images of 62 classes for different land and building types.
The ID and OOD datasets differ in the time of their collection and location, i.e., continent.
We used the standard splits for training, validation, and ID and OOD testing~\cite{Koh21}.
We report the ID testing accuracy and the worst-region OOD testing accuracy, following \citet{Goyal23}.

\noindent
\textbf{Zero-shot} performance is evaluated using ten image classification datasets that cover various image domains: Caltech-101~\cite{Fei-Fei04}, Flowers~\cite{Nilsback08}, Food~\cite{Bossard14}, SUN397~\cite{Xiao10}, DTD~\cite{Cimpoi14}, Aircraft~\cite{Maji13}, StanfordCars~\cite{Krause13}, OxfordPets~\cite{Parkhi12}, EuroSAT~\cite{Helber19}, and UCF-101~\cite{Soomro12}.
We followed the zero-shot evaluation strategy by \cite{Radford21} using prompt engineering.

We used CC3M~\cite{Sharma18_cc3m} as the reference image-caption dataset $\mathcal{S}^{\rm ref}$ to ensure fair comparison with ARF, which also uses this dataset.
We also used Flickr8K~\cite{Rashtchian10} and COCO Captions~\cite{Lin14} for the analysis in Sec.~\ref{sec:anal}.

\subsection{Baselines and Implementation Details}
\label{subsec:baseline}
We considered three key baselines: FLYP~\cite{Goyal23}, ARF~\cite{Han24}, and SnD~\cite{Yu24}.
The first two are robust fine-tuning methods that show promising performance, as discussed in Sec.~\ref{sect:relate}.
SnD was originally designed for continual learning, but we adapted it to our setting.
Specifically, during fine-tuning, it constrains image embeddings as $f_{\theta_{\rm ft}}(\bm{x}^{\rm ref}_j) = f_{\theta_{\rm pre}}(\bm{x}^{\rm ref}_j)$ using a reference dataset.
More details are described in Sec.~\ref{sec:app_snd} in the Appendix.
We also report the performance of the pre-trained model, vanilla FT~\cite{Wortsman22}, and LP-FT~\cite{Kumar22} for a comprehensive comparison.
Both vanilla FT and LP-FT update only the image encoder using the cross-entropy loss with an added classification head.

We used CLIP ViT-B/16~\cite{Dosovitskiy21} pre-trained on LAION-400M~\cite{Schuhmann21} as the base pre-trained model and also used CLIP ViT-L/14 in Table~\ref{tab:res_vitl}.
We report the average results across three seeds.
We reused the codebase and most of the hyper-parameters from FLYP~\cite{Goyal23}, including the AdamW optimizer~\cite{Loshchilov18}, cosine learning rate scheduler, and a batch size of $B=512$ for ImageNet and $B=256$ for other datasets.
Unless otherwise noted, we set $B'=B$.
We also carried out early stopping on the basis of the ID validation performance.
Section~\ref{sec:app_hypara} in the Appendix describes the hyper-parameters and their selection in detail.

Regarding the hyper-parameters in DiVE, we set $\lambda = 1000$ chosen from $\{100, 500, 1000, 2500, 5000\}$ by tuning on the ImageNet validation set and used it for all datasets.
We set $\alpha=0.99$ in Eq.~(\ref{formula:target}); its effect is analyzed in Sec.~\ref{subsec:ablation}.

\subsection{Effect on Preserving Geometric Structure}
We first evaluated the effect of DiVE and the key baselines on preserving the geometric structure of the embeddings, a key aspect for our research question.
We used the representation similarity analysis~(RSA) correlation score~\cite{Kriegeskorte08,Dwivedi19}.
This score quantifies the similarity between all pairwise relationships of embeddings obtained from two different models.
Specifically, we computed the Pearson correlation between the pairwise cosine distance matrices of embeddings from the pre-trained model and from each fine-tuned model, using the Flickr8K dataset.
A higher score indicates better preservation of the geometric structure of the embeddings extracted from the pre-trained model.
Section~\ref{sec:app_rsa} in the Appendix provides the detailed derivation of this score.

Table~\ref{tab:rsa} lists the RSA correlation scores obtained with the baselines and DiVE, including its ablated variants, when using ImageNet as a target task.
Since the official implementation of ARF is not publicly available, we used FLYP with an auxiliary contrastive loss on CC3M as a proxy for ARF, which we refer to as ``FLYP + replay''.
Its implementation details are explained in Sec.~\ref{sec:app_arf} in the Appendix.
We found that the baseline methods struggled to fully preserve the geometric structure of the embeddings extracted from the pre-trained model, as indicated by the RSA correlation scores remaining in 0.825 to 0.850.
In contrast, DiVE and its variants~(with AVL or PVL only) significantly improved the preservation, achieving scores of about 0.98.
These results indicate that DiVE successfully preserves the geometric structure, as intended.
DiVE also achieved a higher score than either AVL or PVL alone, demonstrating the effectiveness of combining these complementary losses.

\begin{table*}[t]
    \centering
        \scalebox{0.96}{
    \begin{tabular}{c c c c c c c c c c c | c}
        \thickhline
          Method           & Caltech & Flowers & Food & SUN & DTD & Aircraft & Cars & Pets & EuroSAT & UCF & ZS avg. \\ \hline
          Pre-trained     & 88.0 & 71.0 & 88.5 & 65.3 & 44.7 & 24.3 & 64.7 & 89.0 & 55.3 & 68.7 & 66.0 \\ \hdashline
          Vanilla FT       & 78.8 & 16.0 & 37.3 & 39.3 & 29.7 & 4.7  & 10.8 & 80.2 & 15.4 & 44.3 & 35.7 \\
          LP-FT            & 84.0 & 44.3 & 68.8 & 49.9 & 37.9 & 15.8 & 37.7 & 81.9 & 30.4 & 59.5 & 51.0 \\
          FLYP$^\dagger$   & 87.6 & 39.7 & 63.3 & 52.6 & 36.8 & 8.0  & 32.3 & 77.2 & 38.2 & 59.0 & 49.5 \\
          ARF              & 88.6 & 46.4 & 74.5 & 63.8 & 40.4 & 13.9 & 44.7 & 83.1 & 35.8 & 64.6 & 55.6 \\ 
          SnD$^\dagger$    & \textbf{89.1} & 49.5 & 69.6 & 58.7 & 38.7 & 11.0 & 42.5 & 79.6 & 42.7 & 62.6 & 54.4 \\
          DiVE~(ours)      & 88.4 & \textbf{66.0} & \textbf{84.3} & \textbf{64.7} & \textbf{47.0} & \textbf{22.1} & \textbf{55.5} & \textbf{88.4} & \textbf{51.4} & \textbf{68.9} & \textbf{63.7} \\
    \thickhline
    \end{tabular}
}
    \vspace{-1.0mm}
    \caption{Zero-shot performance on ten evaluation datasets using CLIP ViT-B/16 on ImageNet.
    Numbers represent top-1 accuracy.
    $\dagger$ denotes our reproduction results.
    Bold indicates best result among fine-tuned methods, i.e., excluding pre-trained model.
    }
    \vspace{-1.0mm}
    \label{tab:zero}
\end{table*}

\subsection{Performance Comparison with Baselines}
\label{subsec:comp}

We evaluated DiVE on ImageNet as a target task and compared its performance with the baselines.
Table~\ref{tab:comp} reports ID and OOD performance, while Table~\ref{tab:zero} presents zero-shot performance.
From Table~\ref{tab:comp}, DiVE achieved the second-best ID performance, despite introducing additional constraints, AVL and PVL, into FLYP.
For OOD performance, DiVE achieved state-of-the-art results on four out of five OOD datasets, as well as the highest average performance across them.
For zero-shot performance in Table~\ref{tab:zero}, DiVE achieved state-of-the-art performance on nine out of ten evaluation datasets, with an average performance of 63.7\%.
This represents a significant improvement over the previous best method, ARF, which achieved 55.6\%.
It is worth noting that the zero-shot performance dropped by only 2.3 points from the pre-trained CLIP ViT-B/16 while improving by 14.2 and 4.5 points in ID and OOD performance, respectively.
As a result, DiVE achieved a well-balanced performance across the three evaluation metrics~(ID, OOD, and zero-shot) at a high level, as shown in Fig.~\ref{fig:radar}~(b).
Given that DiVE significantly improves the preservation of the geometric structure of the embeddings learned by pre-training, as shown in Table~\ref{tab:rsa}, these results support that preserving the geometric structure is effective for robust fine-tuning, providing a positive answer to our research question.
We also confirmed statistical significance on the basis of the results from three seeds, as reported in Sec.~\ref{sec:app_test} in the Appendix.

\begin{table}[t]
    \centering
        \scalebox{0.88}{
    \begin{tabular}{c c c c c c c c c c c c }
        \toprule
                                  &  \multicolumn{3}{c}{iWildCam} & & \multicolumn{3}{c}{FMoW} \\ \cline{2-4} \cline{6-8}
          Method                  & ID   & OOD  & ZS   & & ID   & OOD  & ZS   \\ \midrule
          Pre-trained             & 8.7  & 11.2 & 66.0 & & 20.4 & 18.7 & 66.0 \\ \hdashline
          Vanilla FT              & 48.1 & 35.0 & 50.2 & & 68.5 & 39.2 & 44.6 \\
          LP-FT                   & 49.7 & 34.7 & 51.6 & & 68.4 & 40.4 & 57.9 \\
          FLYP                    & 52.2 & 35.6 & 51.0 & & 68.6 & 41.3 & 45.1 \\ 
          FLYP + replay$^\dagger$ & 48.5 & 35.8 & 62.3 & & 68.7 & 41.2 & 63.0 \\ 
          SnD$^\dagger$           & 50.6 & 37.0 & 60.4 & & 67.0 & 41.4 & 56.6 \\ 
          DiVE                    & \textbf{53.1} & \textbf{37.2} & \textbf{65.3} & & \textbf{69.9} & \textbf{42.3} & \textbf{65.1}\\
    \bottomrule
    \end{tabular}
}
    \vspace{-1.0mm}
    \caption{Results on iWildCam and FMoW.
    ZS indicates average zero-shot performance across ten evaluation datasets.
    Following common practice, we report macro F1-score for iWildCam and top-1 accuracy for FMoW.
    $\dagger$ denotes our reproduction results.
    Bold indicates best result among fine-tuned methods, i.e., excluding pre-trained model.
    }
    \vspace{-1.0mm}
    \label{tab:iwild_fmow}
\end{table}

Table~\ref{tab:iwild_fmow} also shows the results when iWildCam and FMoW are each used as a target task.
We report the average zero-shot performance over ten evaluation datasets.
Since ARF's performance on iWildCam and FMoW is not reported, we used ``FLYP + replay'' as the proxy for ARF.
Consistent with the results on ImageNet, DiVE demonstrated state-of-the-art performance across the three evaluation metrics.
These results indicate that the effectiveness of DiVE is not specific to a particular dataset.

As described in Sec.~\ref{sect:relate}, SnD shares certain concepts with DiVE in that it constrains the difference vectors to be zero.
However, from Tables~\ref{tab:comp}, \ref{tab:zero}, and \ref{tab:iwild_fmow}, SnD consistently underperformed compared to DiVE.
This suggests that constraining the difference vectors to be zero imposes overly strong constraints, which may conflict with the objective of fine-tuning, i.e., adapting to the target data.
We also found that it does not sufficiently preserve the geometric structure of the embeddings, as shown in Table~\ref{tab:rsa}\footnote{We used the same hyper-parameters as used by \cite{Yu24}. While increasing the effect of the constraint by tuning the hyper-parameter improves the RSA correlation score and zero-shot performance, we empirically observed that it severely degrades ID performance, indicating that the fine-tuning process is impeded.}.
These findings suggest that allowing non-zero difference vectors, as DiVE does, provides greater flexibility during fine-tuning.

\begin{table}[t]
    \centering
        \scalebox{1.0}{
    \begin{tabular}{c c | c c | c c c c c c c c }
        \thickhline
          \multicolumn{2}{c|}{Method}   & \multicolumn{2}{c|}{ImageNet} &  \multirow{2}{*}{ZS avg.}  \\ 
          AVL          & PVL            & ID   & OOD avg.               &    \\ \hline
          \multicolumn{2}{c|}{Baseline~(FLYP)} & 82.2 & 59.5                  & 49.5         \\ \hdashline
          $\checkmark$ &                & 82.4 & 62.9~(+3.4)           & 62.9~(+13.4) \\
                       & $\checkmark$   & 82.4 & 62.6~(+3.1)           & 62.7~(+13.2) \\
          $\checkmark$ & $\checkmark$   & 82.5 & 63.2~(+3.7)           & 63.7~(+14.2) \\
    \thickhline
    \end{tabular}
}
    \vspace{-1.0mm}
    \caption{Ablation study of AVL and PVL in DiVE on ImageNet.
    Baseline uses only contrastive loss~(i.e., FLYP).}
    \vspace{-1.0mm}
    \label{table:ab_results}
\end{table}

\begin{table}[t]
    \centering
        \scalebox{1.0}{
    \begin{tabular}{c c c c c c c c c c c c }
    \toprule
     $\alpha$            & ID   & OOD avg. & ZS avg. \\ \midrule
          0              & 82.4 & 61.7     & 61.4 \\ 
          0.5            & 82.5 & 62.6     & 62.4 \\
          0.9            & 82.4 & 62.8     & 62.9 \\
          0.99~(default) & 82.4 & 62.9     & 62.9 \\
    \bottomrule
    \end{tabular}
}
    \vspace{-1.0mm}
    \caption{Effect of $\alpha$ of AVL on performance.}
    \vspace{-1.0mm}
    \label{table:ab_alpha}
\end{table}

\subsection{Ablation Study}
\label{subsec:ablation}
We conducted an ablation study to analyze the effects of AVL and PVL on DiVE's performance.
Table~\ref{table:ab_results} presents the results of applying AVL and PVL individually when using ImageNet as a target task.
AVL significantly improved performance in OOD and zero-shot settings by 3.4 and 13.4 points, respectively.
PVL also yielded improvements in these metrics, though to a lesser extent.
This is because PVL focuses solely on constraining the difference vectors for paired image-caption data samples without ensuring consistency across all difference vectors.
However, when PVL is combined with AVL, it preserves the local geometric structure, thus complementing AVL and enhancing performance.
These results indicate the effectiveness of complementary AVL and PVL, consistent with the results shown in Table~\ref{tab:rsa}.

We also investigated the effect of $\alpha$ in Eq.~(\ref{formula:target}).
Table~\ref{table:ab_alpha} lists the results for various $\alpha$ in AVL.
When $\alpha = 0$, both the OOD and zero-shot performance remained sub-optimal.
This may be because the average vector $\bm{m}$ varies across batches, hindering the effective preservation of the geometric structure.
Increasing $\alpha$ leads to improved performance.
In particular, using a large $\alpha$~(e.g., 0.9 or 0.99) achieved high OOD and zero-shot performance.
These results suggest that AVL enhances OOD and zero-shot performance by providing a stable average vector across batches.

\section{Detailed Analysis}
\label{sec:anal}
This section provides detailed analyses of DiVE.
We conducted experiments on ImageNet as a target task.

\begin{table}[t]
    \centering
        \scalebox{1.0}{
    \begin{tabular}{c c c c c c c c c c c c }
        \toprule
          Method & ID   & OOD avg. & ZS avg. & RSA    \\ \midrule
          Cosine & 82.4 & 62.3     & 61.7    & 0.949  \\
          DiVE   & \textbf{82.5} & \textbf{63.2} & \textbf{63.7} & \textbf{0.981}  \\
    \bottomrule
    \end{tabular}
}
    \vspace{-1.0mm}
    \caption{Comparison between constraints on cosine similarities and difference vectors in DiVE.}
    \vspace{-1.0mm}
    \label{table:cyclip}
\end{table}

\noindent
\textbf{Comparison to Cosine Similarity-based Constraint.}
As described in Sec.~\ref{sect:relate}, CyCLIP~\cite{Goel22} uses the cosine similarity-based constraint to learn a desired geometric structure.
In contrast, DiVE uses the difference vector-based constraint.
To compare these approaches, we modified DiVE to constrain the cosine similarities between embeddings instead of its original vector-based constraint.
Specifically, we replaced the difference vectors, $u(\cdot)$ and $v(\cdot)$, in Eqs.~(\ref{formula:target}), (\ref{formula:tam}), and (\ref{formula:pam}) with $u^{\rm cos}(\cdot)$ and $v^{\rm cos}(\cdot)$.
They are defined as
\begin{align}
\centering
    u^{\rm cos}(\bm{x}^{\rm ref}_j) &= f_{\theta_{\rm ft}}(\bm{x}^{\rm ref}_j) \cdot f_{\theta_{\rm pre}}(\bm{x}^{\rm ref}_j), \\
    v^{\rm cos}(\bm{t}^{\rm ref}_j) &= g_{\phi_{\rm ft}}(\bm{t}^{\rm ref}_j) \cdot g_{\phi_{\rm pre}}(\bm{t}^{\rm ref}_j).
\label{formula:mv_cos}
\end{align}
Table~\ref{table:cyclip} shows the comparison results, where ``Cosine'' indicates the use of $u^{\rm cos}(\cdot)$ and $v^{\rm cos}(\cdot)$.
Constraining cosine similarities, as done with CyCLIP, resulted in a lower RSA correlation score compared with constraining the difference vectors.
It also performed poorly across all evaluation metrics.
This is likely because cosine similarity captures angular closeness on hyperspherical embeddings but not the direction of their difference.
This result highlights the effectiveness of DiVE, which constrains the difference vectors.

\begin{table}[t]
    \centering
        \scalebox{1.0}{
    \begin{tabular}{c c c c c c c c c c c c }
        \toprule
          Method        & ID   & OOD avg. & ZS avg.\\ \midrule
          Pre-training  & 75.2 & 70.9     & 72.7  \\ \hdashline
          FLYP          & 86.0 & 71.5     & 55.4 \\ 
          FLYP + replay & 85.8 & 72.6     & 65.9\\ 
          SnD           & 86.0 & 73.2     & 61.2 \\ 
          DiVE          & \textbf{86.1} & \textbf{74.5} & \textbf{70.1} \\
    \bottomrule
    \end{tabular}
}
    \vspace{-1.0mm}
    \caption{Experimental results on ImageNet using CLIP ViT-L/14.
    Bold indicates best result among fine-tuned methods, i.e., excluding pre-trained model.}
    \vspace{-1.0mm}
    \label{tab:res_vitl}
\end{table}

\begin{table}[t]
    \centering
        \scalebox{0.98}{
    \begin{tabular}{c c c c c c c c c c c c }
        \toprule
          Reference dataset & Size & ID   & OOD avg. & ZS avg. \\ \midrule
          None~(FLYP)       & 0    & 82.2 & 59.5     & 49.5 \\ \hdashline
          Flickr8K          & 8K   & 82.2 & 60.6     & 55.2  \\ 
          COCO Captions     & 118K & 82.4 & 62.3     & 62.2 \\ 
          CC3M~(defalut)    & 3M   & 82.5 & 63.2     & 63.7 \\ 
    \bottomrule
    \end{tabular}
}
    \vspace{-1.0mm}
    \caption{Results of DiVE with various reference datasets.}
    \vspace{-1.0mm}
    \label{tab:ref_dataset}
\end{table}

\noindent
\textbf{Results using ViT-L/14.}
In Sec.~\ref{sect:exp}, we used CLIP ViT-B/16 as the base model.
To evaluate the generality across architectures, we report results using CLIP ViT-L/14 pre-trained on LAION-400M in Table~\ref{tab:res_vitl}.
For methods that use the reference dataset~(CC3M), i.e., FLYP + replay, SnD, and DiVE, we decreased $B'$ to 256 due to GPU memory constraints.
Details of our computing infrastructure are provided in Sec.~\ref{sec:app_infra} in the Appendix.
We observe that similar trends hold for ViT-L/14 as with ViT-B/16, i.e., DiVE achieved strong results across the three metrics.
This indicates that the effectiveness of DiVE is not tied to a specific architecture and scales well to larger models, such as ViT-L/14.

\noindent
\textbf{Reference Datasets.}
We evaluated the impact of the choice of the reference dataset.
Table~\ref{tab:ref_dataset} presents the results of DiVE with Flickr8K, COCO Captions, and CC3M as the reference dataset.
We confirm that all variations consistently outperformed the baseline, which does not use a reference dataset~(i.e., FLYP).
We also observed that performance improved as the dataset size increased.
This trend may be because larger datasets cover a wider region in the embedding space. 
Despite this trend, it is worth noting that, even when using COCO Captions, DiVE improved both OOD and zero-shot performance over ARF~(relying on CC3M).

\section{Conclusion}
We proposed Difference Vector Equalization~(DiVE), a robust fine-tuning method for contrastive pre-trained vision-language models.
DiVE aims to preserve the geometric structure of the embeddings learned by pre-training, enabling fine-tuning without compromising the generalization ability.
It thus constrains the difference vectors caused by fine-tuning to be equal across various data samples.
We introduced two losses: average vector loss~(AVL) and pairwise vector loss~(PVL).
AVL preserves the geometric structure globally by constraining difference vectors to be equal to their weighted average.
PVL preserves the geometric structure locally by ensuring a consistent image-caption alignment.
Extensive experiments demonstrated that DiVE significantly improves the performance of baseline methods by preserving the geometric structure.
We believe that DiVE is not only a practical method but also opens a new direction in fine-tuning vision-language models by offering strong evidence for the importance of preserving the geometric structure of the embeddings.
As future work, we aim to deepen the theoretical understanding of DiVE and clarify the principles underlying its effectiveness.

\bibliography{suzuki-aaai}

\setcounter{secnumdepth}{2}

\clearpage
\setcounter{page}{1}
\appendix

\section{Integration into Weight Ensemble}
\label{sec:app_wise}
Ensembling the weights of pre-trained and fine-tuned models improves generalization ability, as described in Sec.~\ref{sect:relate}.
We integrated DiVE into this strategy.
Following previous studies~\cite{Wortsman22,Ilharco22}, we applied a simple linear interpolation between the weights of the pre-trained and fine-tuned models.
The interpolation coefficient was selected from $\{0.1, 0.2,\cdots, 0.9\}$ to maximize ID validation performance, following \citet{Goyal23}.

Table~\ref{table:app_weight_ens} shows the comparison between DiVE and its variant with weight ensembling when using ImageNet as a target task.
We observed that integrating DiVE into the weight ensembling strategy improved performance across all three evaluation metrics.
Both DiVE and its ensembling version outperformed FLYP with weight ensembling.

\begin{table}[ht]
    \centering
        \scalebox{1.0}{
    \begin{tabular}{c c c c c c c c c c c c }
        \toprule
          Method          & ID   & OOD avg. & ZS avg.\\ \midrule
          FLYP + ensemble & 82.5 & 62.3     & 57.3   \\ \hdashline
          DiVE            & 82.5 & 63.2     & 63.7   \\
          DiVE + ensemble & \textbf{82.6} & \textbf{63.5}     & \textbf{64.6}   \\
    \bottomrule
    \end{tabular}
}
    \vspace{-1.0mm}
    \caption{Comparison between FLYP with weight ensemble, DiVE, and its variant with weight ensemble.}
    \vspace{-1.0mm}
    \label{table:app_weight_ens}
\end{table}

\section{Details of SnD}
\label{sec:app_snd}
As described in Sec.~\ref{sect:relate}, SnD~\cite{Yu24} was originally designed for continual learning, but it shares certain concepts with our DiVE.
Therefore, we used it as a key baseline.
Similar to DiVE, SnD uses an additional reference dataset $\mathcal{S}^{\rm ref}$ to constrain the image embeddings during continual learning.
Specifically, it constrains the image embeddings from the pre-trained and fine-tuning models to be identical by using the loss defined as
\begin{align}
\centering
    \mathcal{L}_{\rm snd} = \frac{1}{B'}\sum^{B'}_{j=1} \| f_{\theta_{\rm ft}}(\bm{x}^{\rm ref}_j) - f_{\theta_{\rm pre}}(\bm{x}^{\rm ref}_j) \|^2. 
\label{formula:app_snd}
\end{align}
In our experiments, we combined this loss into the contrastive loss $\mathcal{L}_{\rm cl}$.
The loss function for fine-tuning the pre-trained model is defined as
\begin{align}
\centering
    \mathcal{L}_{\rm final\_snd} &= \mathcal{L}_{\rm cl} + \lambda_{\rm snd} \cdot \mathcal{L}_{\rm snd},
\label{formula:app_snd_final}
\end{align}
where $\lambda_{\rm snd}$ is a hyper-parameter that controls the strength of $\mathcal{L}_{\rm snd}$.
As shown in Eq.~(\ref{formula:app_snd}), SnD only constrains the image embeddings because the original SnD minimizes the cross-entropy and updates only the image encoder.
In contrast, the formulation in Eq.~(\ref{formula:app_snd_final}) also updates the text encoder, enabling the introduction of the constraint for text embeddings.
However, we empirically found that this approach decreases performance.
Therefore, we adopt the original formulation shown in Eq.~(\ref{formula:app_snd}).

In the implementation of SnD, we used CC3M as the reference dataset for fair comparison with DiVE.
We also set $\lambda_{\rm snd}=1$ following \citet{Yu24}.

\section{Details of Hyper-parameters}
\label{sec:app_hypara}

As described in Sec.~\ref{subsec:baseline}, we mainly followed the hyper-parameter settings of FLYP~\cite{Goyal23} across all methods in our experiments.
For the learning rate, we selected the optimal value for each method from $\{3\times10^{-4}, 1\times10^{-4}, 3\times10^{-5}, 1\times10^{-5}, 3\times10^{-6}\}$ on the basis of the performance on each validation set.
Table~\ref{table:app_hypara} lists the hyper-parameters used in DiVE.
For other methods in our experiments, the selected learning rate was consistently $1\times10^{-5}$ across all three datasets.
All hyper-parameters except the learning rate were the same as those listed in Table~\ref{table:app_hypara}.

\begin{table}[ht]
    \centering
        \scalebox{1.0}{
    \begin{tabular}{l c c c c c c c c c c c }
        \toprule
          Configuration      & ImageNet & iWildCam & FMoW   \\ \midrule
          Optimizer          & AdamW    & AdamW    & AdamW  \\
          Batch size~($B$)   & 512      & 256      & 256    \\
          Weight decay       & 0.1      & 0.2      & 0.2    \\
          Warm-up steps      & 500      & 500      & 500    \\
          Learning rate~(LR) & $1\times10^{-5}$ & $3\times10^{-5}$ & $3\times10^{-5}$ \\
          LR schedule        & Cosine   & Cosine   & Cosine \\
          Total epochs       & 10       & 20       & 20     \\
    \bottomrule
    \end{tabular}
}
    \vspace{-1.0mm}
    \caption{Hyper-parameters for optimization used in DiVE.}
    \vspace{-1.0mm}
    \label{table:app_hypara}
\end{table}

\section{Derivation of RSA Correlation Score}
\label{sec:app_rsa}

We used the representation similarity analysis~(RSA)~\cite{Kriegeskorte08} to evaluate the effect on preserving the geometric structure of the embeddings.
The RSA correlation score is widely used to compare a computational or behavioral model with brain responses.
This score has also been used to compare embeddings from deep neural networks and evaluate their similarity~\cite{Dwivedi19}.
We explain the derivation of this score in our evaluation.

We first consider two sets of embeddings, $\{\bm{a}_k\}^{M}_{k=1}$ and $\{\bm{b}_k\}^{M}_{k=1}$, each consisting of $M$ embeddings with dimension $d$.
We then compute a representation dissimilarity matrix~(RDM) for each set.
The RDM is an $M \times M$ matrix, where each entry $(i, j)$ measures the dissimilarity between the $i$-th and $j$-th embeddings within the set.
Although arbitrary metrics can be used to compute dissimilarity~(e.g., Euclidean distance), we used cosine distance in our experiments.
Therefore, each element of the RDM is defined as
\begin{align}
\centering
    \bm{D}^1_{i,j} = 1 - {\rm cos}(\bm{a}_i,\bm{a}_j),\ \bm{D}^2_{i,j} = 1- {\rm cos}(\bm{b}_i,\bm{b}_j),
\label{formula:app_rdm}
\end{align}
where $\bm{D}^1$ and $\bm{D}^2$ denote the RDMs computed from the embedding sets $\bm{a}$ and $\bm{b}$, respectively.
Furthermore, ${\rm cos}(\cdot,\cdot)$ calculates the cosine similarity between two embeddings.
Since $L_2$-normalized embeddings are used in our experiments, this can be achieved by a simple inner product.

Finally, we vectorize the upper triangular parts~(excluding the diagonal) of both RDMs and compute the Pearson correlation coefficient between them.
This score ranges from $-1$ to $1$, with higher values indicating better preservation of the geometric structure in the embedding space between the two models~(e.g., pre-trained and fine-tuned).

For our evaluation, we used Flickr8K~\cite{Rashtchian10} as the image-caption dataset.
We extracted image and text embeddings from the pre-trained model~(used as $\bm{a}$) and fine-tuned model~(used as $\bm{b}$), respectively.

\section{Details of Proxy for ARF}
\label{sec:app_arf}

As described in Sec.~\ref{subsec:baseline}, ARF serves as a key baseline for DiVE.
However, the official implementation of ARF is not publicly available.
Therefore, we used a proxy for ARF~(called ``FLYP + replay'') to report the results shown in Tables~\ref{tab:rsa}, \ref{tab:iwild_fmow}, \ref{tab:res_vitl}, \ref{table:app_std}, and \ref{table:app_pvalue}.
In FLYP + replay, we introduce the auxiliary contrastive loss $\mathcal{L}_{\rm aux}$ into FLYP.
The $\mathcal{L}_{\rm aux}$ calculates the contrastive loss defined in Eq.~(\ref{formula:cl}) on the reference dataset $\mathcal{S}^{\rm ref}$.
The loss function for fine-tuning the pre-trained model is defined as
\begin{align}
\centering
    \mathcal{L}_{\rm final\_aux} = \mathcal{L}_{\rm cl} + \lambda_{\rm aux}  \cdot \mathcal{L}_{\rm aux}.
\label{formula:proxy}
\end{align}
where $\lambda_{\rm aux}$ is a hyper-parameter that controls the strength of $\mathcal{L}_{\rm aux}$.
Following \citet{Han24}, we used CC3M as $\mathcal{S}^{\rm ref}$ and $\lambda_{\rm aux} = 1$.
FLYP + replay achieved higher OOD and zero-shot performance than ARF~(see Tables~\ref{tab:comp}, \ref{tab:zero}, and \ref{table:app_std}), indicating that it serves as a suitable proxy.

\section{Statistical Analysis of Performance Gaps}
\label{sec:app_test}

We statistically analyzed the performance gap between DiVE and key baselines using the results from three runs with different random seeds.
We used ``FLYP + replay'' as the proxy for ARF.
Table~\ref{table:app_std} summarizes the ID, OOD, and zero-shot performance on ImageNet, including standard deviations.
The standard deviations are small, indicating the results are consistent across all three runs.
Table~\ref{table:app_pvalue} reports the $p$-values from paired $t$-tests comparing DiVE with each baseline across ID, OOD, and zero-shot metrics on ImageNet.
The effect of DiVE was statistically significant at a high confidence level in all cases, except for its comparison with SnD on ID performance.
These findings support the reliability of improvements in DiVE.

\begin{table}[ht]
    \centering
        \scalebox{1.0}{
    \begin{tabular}{c c c c c c c c c c c c }
        \toprule
          Method        & ID                          & OOD avg.                    & ZS avg.  \\ \midrule
          FLYP          & 82.2 \scriptsize{$\pm$ 0.0} & 59.5 \scriptsize{$\pm$ 0.1} & 49.5 \scriptsize{$\pm$ 0.2}\\
          FLYP + replay & 82.1 \scriptsize{$\pm$ 0.1} & 61.4 \scriptsize{$\pm$ 0.1} & 58.1 \scriptsize{$\pm$ 0.3}\\
          SnD           & 82.4 \scriptsize{$\pm$ 0.1} & 60.7 \scriptsize{$\pm$ 0.2} & 54.4 \scriptsize{$\pm$ 0.1}\\
          DiVE          & 82.5 \scriptsize{$\pm$ 0.0} & 63.2 \scriptsize{$\pm$ 0.1} & 63.7 \scriptsize{$\pm$ 0.1}\\
    \bottomrule
    \end{tabular}
}
    \vspace{-1.0mm}
    \caption{ID, OOD, and zero-shot performance on ImageNet, with standard deviations.}
    \vspace{-1.0mm}
    \label{table:app_std}
\end{table}

\begin{table}[ht]
    \centering
        \scalebox{0.9}{
    \begin{tabular}{c c c c c c c c c c }
        \toprule
          Baseline          & ID                            & OOD avg.                      & ZS avg.  \\ \midrule
          vs. FLYP          & $\mathbf{5.5 \times 10^{-3}}$ & $\mathbf{4.5 \times 10^{-4}}$ & $\mathbf{1.2 \times 10^{-4}}$ \\
          vs. FLYP + replay & $\mathbf{8.8 \times 10^{-3}}$ & $\mathbf{1.5 \times 10^{-3}}$ & $\mathbf{1.0 \times 10^{-3}}$ \\
          vs. SnD           & $2.6 \times 10^{-1}$          & $\mathbf{4.4 \times 10^{-4}}$ & $\mathbf{5.2 \times 10^{-5}}$ \\
    \bottomrule
    \end{tabular}
}
    \vspace{-1.0mm}
    \caption{$p$-values of paired $t$-tests comparing DiVE against each baseline on ImageNet.
    Bold indicates statistically significant result where null hypothesis of paired $t$-test was rejected at $p < 5.0 \times 10^{-2}$.
    }
    \vspace{-1.0mm}
    \label{table:app_pvalue}
\end{table}

\section{Computing Infrastructure and Computational Costs}
\label{sec:app_infra}

In all our experiments, we used a machine equipped with eight NVIDIA H200 GPUs, each with 140 GB of memory.
For the software, we reused the codebase from FLYP~\cite{Goyal23}, which is built on the PyTorch framework.
We used Ubuntu 24.04 as the operating system.

Under these infrastructures, Table~\ref{table:app_comp_cost} compares the computational costs of FLYP and DiVE.
As shown, DiVE increases the memory usage and the training time.
This is because FLYP requires only a single forward-backward pass, whereas DiVE requires extra passes over the reference dataset.
Although the implementation details of ARF are not publicly available, we expect it to exhibit a similar trend.
This additional cost may become a limitation when scaling to larger models.
Therefore, mitigating the computational overhead introduced by the reference dataset constitutes an important direction for future research.

\begin{table}[ht]
    \centering
        \scalebox{0.9}{
    \begin{tabular}{c c c c c c}
        \toprule
          Method  & Memory~[MiB] & Training time~[min/epoch] \\ \midrule
          FLYP    & 116,972      & 35.9 \\ 
          DiVE    & 321,254      & 58.9 \\ 
    \bottomrule
    \end{tabular}
}
\vspace{-2.0mm}
    \caption{Computational costs on ViT-B/16 when using ImageNet as target task.}
\label{table:app_comp_cost}
\end{table}

\end{document}